\documentclass[10pt,twocolumn,letterpaper]{article}

\usepackage{cvpr}              % To produce the CAMERA-READY version
\usepackage[dvipsnames]{xcolor}

\usepackage{multirow}

\definecolor{cvprblue}{rgb}{0.21,0.49,0.74}
\usepackage[pagebackref,breaklinks,colorlinks,citecolor=cvprblue]{hyperref}

\def\paperID{*****} % *** Enter the Paper ID here
\def\confName{CVPR}
\def\confYear{2024}

\title{The System Description of CPS Team for Track on Driving with Language of CVPR 2024 Autonomous Grand Challenge}

\author{Jinghan Peng\textsuperscript{*}, Jingwen Wang\textsuperscript{*}, Xing Yu\textsuperscript{*}, Dehui Du\textsuperscript{\dag} \\
East China Normal University \\
{\tt\small \{51255902115, 51265902037, 51265902017\}@stu.ecnu.edu.cn, dhdu@sei.ecnu.edu.cn}
}

\begin{document}
\maketitle

\begingroup\renewcommand\thefootnote{*}
\footnotetext{These authors share equal contribution to this work.}
\begingroup\renewcommand\thefootnote{\dag}
\footnotetext{Corresponding author.}

\begin{abstract}
% The ABSTRACT is to be in fully justified italicized text, at the top of the left-hand column, below the author and affiliation information.
% Use the word ``Abstract'' as the title, in 12-point Times, boldface type, centered relative to the column, initially capitalized.
% The abstract is to be in 10-point, single-spaced type.
% Leave two blank lines after the Abstract, then begin the main text.
% Look at previous \confName abstracts to get a feel for style and length.
This report outlines our approach using vision language model systems for the Driving with Language track of the CVPR 2024 Autonomous Grand Challenge. We have exclusively utilized the DriveLM-nuScenes dataset for training our models. Our systems are built on the LLaVA models, which we enhanced through fine-tuning with the LoRA and DoRA methods. Additionally, we have integrated depth information from open-source depth estimation models to enrich the training and inference processes. For inference, particularly with multiple-choice and yes/no questions, we adopted a Chain-of-Thought reasoning approach to improve the accuracy of the results. This comprehensive methodology enabled us to achieve a top score of 0.7799 on the validation set leaderboard, ranking 1st on the leaderboard.

% This report describes our vision language model systems for the track on Driving with Language of the CVPR 2024 Autonomous Grand Challenge. 
% In this project, we exclusively use the DriveLM-nuScenes dataset for training. 
% We utilize LLaVA models as the base models. 
% We fine-tune the base models using the LoRA and DoRA methods.
% % We perform data cleaning and deduplication on the training set to ensure the quality of the training process. 
% Additionally, we incorporate depth information from open-source depth estimation models into the training and inference processes.
% During inference, for multiple-choice and yes/no questions, we employ a Chain-of-Thought reasoning approach to enhance result accuracy. 
% Ultimately, we achieved a final score of 0.7799 on the validation set leaderboard.

\end{abstract}    
\section{Introduction}
\label{sec:intro}

%多模态大语言模型在自动驾驶的应用
Large Language Models (LLMs) and Vision Language Models (VLMs) have achieved remarkable advancements in recent years. 
While traditional LLMs primarily focus on processing and generating natural language, VLMs extend these capabilities by integrating vision data with language, enabling the processing of text and images simultaneously~\cite{ijcai2022p762}\cite{zhang2024vision}.
In autonomous driving, VLMs are particularly advantageous due to their capability to process and integrate visual and language information concurrently.
This capability enables the detection of obstacles, recognition of traffic signs, interpretation of road conditions, and informed decision-making based on these visual cues. 
% Additionally, the integration of language facilitates the processing of voice commands and natural language instructions, enhancing vehicle-driver interaction.
This combination of vision and language processing enhances the overall performance and safety of autonomous driving systems (ADS)~\cite{cui2024survey}. 
In summary, the ongoing development of VLMs holds vast potential for autonomous driving, promising more intelligent and reliable systems.

Building on the progress in VLMs, the Driving with Language track of CVPR 2024 Autonomous Grand Challenge serves as a platform to test and demonstrate how this technology can be adapted for practical applications, highlighting the significance of integrating advanced vision-language models into real-world autonomous driving scenarios.
This track requires participants to develop models that integrate language modality to address complex driving questions by reasoning and making decisions using multi-view image inputs across various driving scenarios.
To address this challenge, we propose a solution aimed at achieving accurate environmental perception, precise motion prediction, and generalizable and explainable driving behavior.
We utilize depth estimation models to obtain depth information from images and construct a high-quality dataset through carefully designed prompts.
We further fine-tune vision-language models using parameter-efficient fine-tuning methods.
These refined models are integrated into a meticulously designed inference pipeline, enhancing the reasoning capabilities by leveraging both vision and language data, thereby significantly improving overall performance.
Finally, we integrate the results from multiple individual systems to achieve further enhancements in performance.

% This report provides a detailed description of our model's architecture and our comprehensive approach. 
The report is structured as follows: \cref{sec:dataset} details the training and validation datasets employed, including the preprocessing methods used. \cref{sec:system} introduces the base model of our VLMs. \cref{sec:training} describes the fine-tuning process and specifics of the VLMs. \cref{sec:inference} describes the inference methodology and the incorporation of depth information. \cref{sec:result} presents experimental results and performance analysis. Finally, \cref{sec:conclusion} provides a summary of our findings and contributions.

%Incorporating the language modality, this task connects Vision Language Models (VLMs) and autonomous driving systems. The model will introduce the reasoning ability of LLMs to make decisions, and pursue generalizable and explainable driving behavior. Given multi-view images as inputs, models are required to answer questions covering various aspects of driving.

\section{Dataset}
\label{sec:dataset}

\subsection{Training Dataset}
\label{sec:trainset}
For the Driving with Language track, we use the DriveLM-nuScenes~\cite{sima2023drivelm} dataset as the training dataset. This dataset consists of 4072 sample frames from 696 scenes in the nuScenes~\cite{nuscenes2019} dataset, comprising a total of 377,983 questions. 
Each scene is composed of a series of sample frames. Each sample frame includes six images, information on several key objects in the current scene, and a series of question-and-answer (QA) pairs. As illustrated in~\cref{fig:images}, the six images, each with a resolution of $1600\times 900$, are taken from six cameras mounted at different directions on the vehicle.
The key object information defines the crucial objects in the current scene, including the status of specified objects, their visual descriptions, and their 2D object bounding boxes in the images.
As shown in~\cref{fig:images}, each key object is identified using a unique \textit{KeyObj Tag}.
These QA pairs encompass multiple-choice, yes/no, and dialogue-based formats, covering perception, prediction, planning, and behavior tasks related to driving.

To enhance the model’s ability to identify key objects accurately, we utilize key object information from the training set to generate additional QA pairs for training. Examples are provided below, where the answer forms the description for the key object.

% \medskip
\noindent
{\bf Q:} \textit{The width and height of the image are 1600 and 900 respectively. $<$c4,CAM\_FRONT,920.8,383.3$>$ represents the key object that the center coordinates of the bounding box in the CAM\_FRONT image are (920.8,383.3). What is the object $<$c4,CAM\_FRONT,920.8,383.3$>$? What is the state of it?}\\
{\bf A:} \textit{$<$c4,CAM\_FRONT,920.8,383.3$>$ is a white truck to the front of the ego vehicle. It is moving.}
% \medskip

To enhance the precision of object depth information, we utilize the open-source depth estimation model called Depth Anything~\cite{yang2024depth} to estimate the depth of images, as depicted in~\cref{fig:depth}. We calculate depth estimates for each pixel within the established bounding boxes of key objects in the training set. From these estimates, we select the 75th percentile value as the representative depth for each object. This depth value is then translated into a textual distance description, such as 'close' or 'far', which is subsequently incorporated into the object’s descriptive metadata.
% For key objects in the test set, we extract the depth values for each pixel within an $11\times11$ rectangular frame centered on the object’s coordinates and calculate the object’s depth estimate using the same method.
% For the two types of multiple-choice questions that appear in the validation set, we convert the questions and answers into the multiple-choice format using the same method as provided in the official script.

\begin{figure*}
  \centering
  \includegraphics[width=0.82\textwidth]{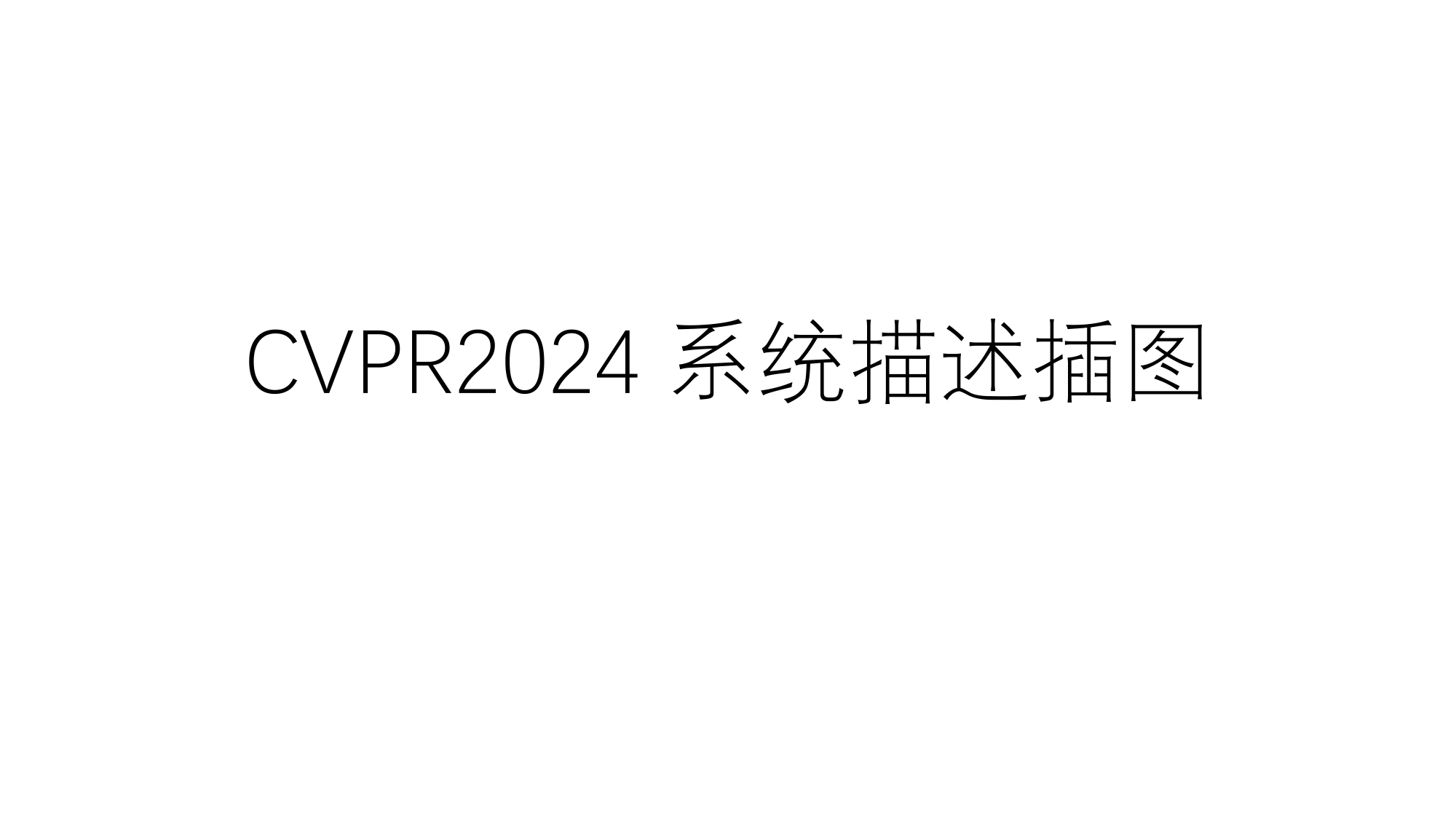}
  \caption{
Diagram of six images from a sample frame, captured by cameras mounted on the vehicle in six directions: front, front left, front right, back, back left, and back right. Key objects are marked with IDs and 2D bounding boxes.
  }
  \label{fig:images}
\end{figure*}

\begin{figure*}
  \centering
  \includegraphics[width=0.82\textwidth]{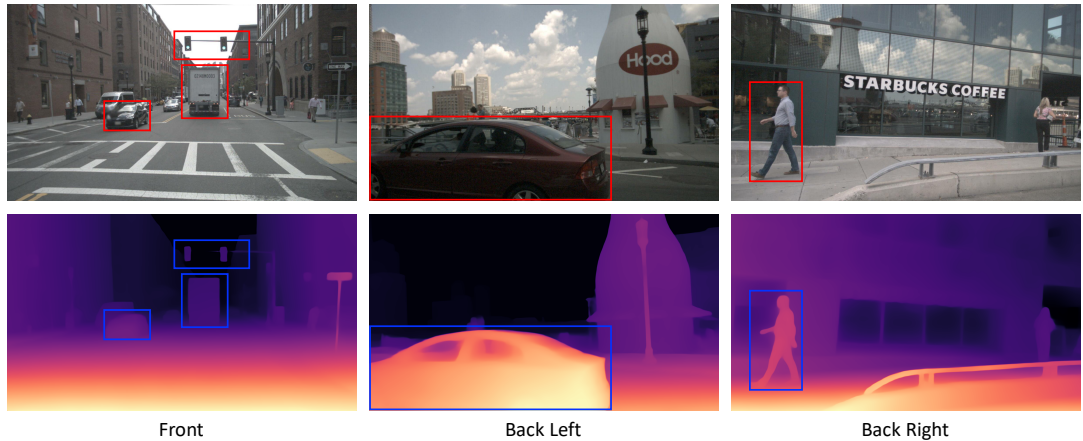}
  \caption{
The diagram displays three raw images (Top) and their corresponding depth estimation images (Bottom) from a sample frame. Key objects are highlighted with red bounding boxes in the raw images and blue bounding boxes in the depth estimation images.
  }
  \label{fig:depth}
\end{figure*}

\subsection{Validation Dataset}
% The evaluation dataset has the same data distribution as the training dataset. 
% The evaluation dataset consists of 799 sample frames from 149 scenes in the nuScenes~\cite{nuscenes2019} dataset, comprising a total of 15,480 questions. 
The validation dataset, which mirrors the data distribution of the training dataset, consists of 799 sample frames drawn from 149 scenes in the nuScenes dataset \cite{nuscenes2019}, encompassing a total of 15,480 questions.
% Similar to the training dataset, these questions include multiple-choice questions, yes/no questions, and dialogue-based questions focusing on tasks associated with perception, prediction, planning, and behavior.
Different questions are evaluated using different metrics, and the final score is calculated by a weighted sum of these various evaluation scores.
% The evaluation dataset does not cover all types of questions that appear in the training dataset.
% For key objects in the validation set, we extract the depth values for each pixel within an $11\times11$ rectangular frame centered on the object’s coordinates and calculate the object’s depth estimate using the same method as in training dataset.
For key objects in the validation set, we extract their coordinates based on the \textit{KeyObj Tag} and retrieve depth values for each pixel within an \(11 \times 11\) rectangular frame centered on these coordinates. We then calculate the depth estimates for these objects using the same method applied in the training dataset.
\section{Systems}
\label{sec:system}
LLaVA (Large Language and Vision Assistant) is an advanced multimodal model that effectively processes both visual and language data by combining a vision encoder with a language model. This integration allows it to perform complex tasks involving both text and images~\cite{liu2023improved}. Our systems utilize LLaVA-1.5-7B and LLaVA-NeXT-7B as foundational models for further fine-tuning.

LLaVA-1.5 integrates a pre-trained visual encoder with the pre-trained language model, allowing for a robust understanding and generation of content across both modalities. The model is particularly innovative in its use of GPT-4 to generate multimodal instruction-following data from image-text pairs, which are then used for instruction tuning on machine-generated data~\cite{liu2024visual}. This approach enables LLaVA-1.5 to handle a wide array of scenarios including conversations, detailed descriptions, and complex reasoning tasks. Its architecture facilitates seamless integration between visual features and textual data through a two-stage training process involving pre-training for feature alignment followed by end-to-end fine-tuning.

Building on the capabilities of LLaVA-1.5, LLaVA-NeXT introduces significant improvements such as enhanced visual reasoning, OCR capabilities, and an expanded understanding of world knowledge. It processes images at higher resolutions and accommodates three different aspect ratios, thus capturing more visual details. 
% Additionally, LLaVA-NeXT employs advanced visual instruction tuning and improved conversation capabilities, drawing from a richer mixture of data sources~\cite{liu2024llavanext}. 
These developments ensure that LLaVA-NeXT significantly advances beyond its predecessor in terms of performance and applicability in real-world applications.

\section{Training Protocol}
\label{sec:training}

We fine-tune the LLaVA model using the training data described in \cref{sec:trainset}. 
To optimize computational and parameter efficiency, we have opted against using a full fine-tuning approach to train our model. 
Instead, we employ LoRA~\cite{hu2021lora}, an efficient parameter fine-tuning method, to fine-tune all fully connected layers within the language model component of LLaVA. 
Furthermore, we explore the use of DoRA~\cite{liu2024dora}, an advanced version of LoRA, to enhance our fine-tuning process.
During the training process, we extract information about the key objects mentioned within each question. We select the corresponding image that contains these key objects to serve as the image input for the model. 
We then prepend the descriptions of these key objects to the question before inputting it into the model. 
For questions that solely contain directional information, we select the corresponding image based on the specified direction as the input. 
We then concatenate descriptions of key objects visible from that direction, prepending this information to the question.
For questions lacking both key object details and directional information, we opt for an image facing forward as the input.
We then concatenate descriptions of all key objects in view and prepend this information to the question.
For all our experiments, we employ PyTorch framework on a computation platform with an Intel Xeon Gold 5218R CPU, eight NVIDIA RTX 3090 GPUs, and 256 GB of memory. 
For the LoRA and DoRA configurations, we set the rank and alpha to 8 and 16, respectively.
We implement a cosine learning rate scheduler starting at an initial rate of 2e-5 and incorporate a warm-up phase during the first 3\% of the training steps.
Each system is fine-tuned on the training set for one epoch to prevent overfitting.

\section{Inference}
\label{sec:inference}
Our proposed inference framework is illustrated in~\cref{fig:infer}. For a single inference, a text-based question and a scene image are processed through a prompt design module that integrates a depth estimation model with our VLM, creating a prompt enriched with detailed scene information. This prompt and the scene image are then input into the VLM to generate the final answer.
\begin{figure*}
  \centering
  \includegraphics[width=1\textwidth]{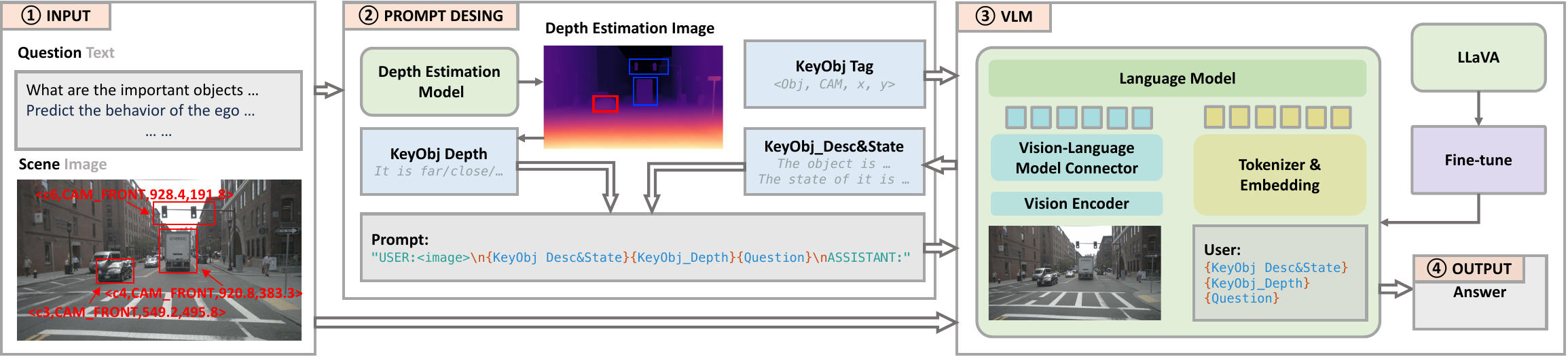}
  \caption{
  The architecture of our proposed inference framework.
  }
  \label{fig:infer}
\end{figure*}

\begin{table*}[]
\resizebox{\textwidth}{!}{
\begin{tabular}{l|ccccccccc|c}
\toprule
\multirow{2}{*}{Systems} & \multirow{2}{*}{Accuracy$\uparrow$} & \multirow{2}{*}{ChatGPT$\uparrow$} & \multicolumn{5}{c}{Language$\uparrow$} & \multicolumn{1}{c}{} & \multirow{2}{*}{Match$\uparrow$} & \multirow{2}{*}{Final Score$\uparrow$} \\
 & \multicolumn{1}{c}{\textbf{}} & \multicolumn{1}{c}{} & \multicolumn{1}{c}{Bleu\_1} & \multicolumn{1}{c}{Bleu\_2} & \multicolumn{1}{c}{Bleu\_3} & \multicolumn{1}{c}{Bleu\_4} & \multicolumn{1}{c}{ROUGE\_L} & \multicolumn{1}{c}{CIDEr} & \multicolumn{1}{c|}{} & \multicolumn{1}{c}{\textbf{}} \\ 
 \midrule
LLaVA-1.5-7B+LoRA & 0.7105 & 59.9425 & 0.9267 & 0.9040 & 0.8822 & 0.8602 & 0.9275 & 7.3282 & 90.3968 & 0.7329 \\
LLaVA-1.5-7B+DoRA & 0.6944 & 61.4596 & 0.8805 & 0.8228 & 0.7728 & 0.7269 & 0.8691 & 5.3107 & 93.1453 & 0.7177 \\
LLaVA-NeXT-7B+LoRA & 0.7431 & 64.5203 &  \textbf{0.9411} & \textbf{0.9212} & \textbf{0.9008} & \textbf{0.8798} & \textbf{0.9377} & \textbf{7.8806} & 93.7031  & 0.7710 \\ 
\midrule
Fusion & \textbf{0.7861} & \textbf{64.8120} & \textbf{0.9411} & \textbf{0.9212} & \textbf{0.9008} & \textbf{0.8798} & \textbf{0.9377} & \textbf{7.8806} & \textbf{93.8406} & \textbf{0.7799} \\ 
\bottomrule
\end{tabular}
}
\caption{The table shows the top results for our individual systems and the fusion system on the validation dataset.}
\label{tab:result}
\end{table*}

During our preliminary testing, we discovered that the accuracy of the responses was relatively low due to insufficient information available from the initial input. To enhance the accuracy of answers, we designed a pipeline to guide the VLM towards more precise reasoning. For the initial input \textit{Question Text}, heuristic rules are applied to extract \textit{KeyObj Tag} of key objects as the off-line label for subsequent use. For the \textit{Scene Image}, we first employ the depth estimation model to compute the depth information of the image. This depth information, combined with the \textit{KeyObj Tag}, is used to estimate the depth of the key objects, \textit{KeyObj Depth}, which serves as one of the components for constructing the subsequent prompt. Additionally, to improve the quality of the response of VLM, we isolate the process of describing and reasoning about the key objects' states. This isolated query to the VLM includes the key objects' descriptions and states in the scene image \textit{KeyObj Desc\&State}, which also forms part of the information for constructing the prompt. The final generated prompt format is as follows:
% 本章将介绍我们提出的inference流程框架，呈现如figure所示。对于一次推理，输入是text格式的question和image格式的场景，经过prompt design模块，结合深度估计模型和VLM处理信息并组合成具有详细场景信息的prompt，再将新的prompt与scene image输入VLM，最终得到text格式的answer。
% 在我们前期的测试过程中，发现由于模型的模态能力限制，回答的准确性较低。为了提升回答的准确性，我们设计了一套pipeline来指导VLM进行更精确的推理。对于initial input的question text，按照启发式规则对其中涉及的key object进行提取，得到的KeyObj Tag留待利用。对于scene image，我们首先用深度估计模型Depth Anything~\cite{yang2024depth}计算图片深度信息，结合KeyObj Tag计算关键物体对应的KeyObj Depth，作为后续prompt构造的信息之一。另一方面，为了提高后续VLM推理和回答的质量，我们将VLM对关键物体的描述和状态的推理过程独立出来，单独将提问VLM，KeyObj Tag对应的关键物体在scene image中的KeyObj Description and State，同样作为后续prompt的构造信息。最后，我们生成的prompt格式如下：

% \medskip
\noindent
{\bf Prompt:} \textit{USER: $<$image$>$ \{KeyObj Desc\&State\}\{KeyObj Depth\}\{Question\} ASSISTANT:}
\medskip

This prompt includes the initial text information of the question, the depth information of the key objects, and the description and state information of the key objects. It is used along with the scene image as the final input to the VLM, from which the model's inference result is derived. 
It is important to note that we employ a multi-system fusion approach, utilizing the best-performing model for each question type to organize the final inference results.

% 这个prompt同时包含了初始的问题文本信息、关键物体的深度信息、关键物体的描述和状态信息。将其和scene image一起作为最终输入给VLM，得到模型的推理结果。需要额外说明的一点是，我们使用multi-system fusion的方法，针对不同的问题类型使用了在该类型上表现最好的模型来组织最后的推理结果。
Chain-of-Thought prompting (CoT)~\cite{wei2022chain} facilitates complex reasoning through intermediate steps. 
Our approach combines Zero-shot CoT Prompting~\cite{kojima2022large} and Few-shot CoT prompting~\cite{brown2020language} to enhance response accuracy for critical question types, including multiple-choice and yes/no questions. Despite our tailored prompt design, we observed a decrease in model performance during evaluations. Further analysis revealed that applying few-shot CoT in the complex context of autonomous driving may have inadvertently constrained the model's inherent reasoning abilities, limiting its capacity to generate diverse responses. Two main strategies for improvement include designing more comprehensive prompts to accommodate a variety of scenarios and fundamentally enhancing the CoT ability of model~\cite{kim2023cot}.

\section{Result}
\label{sec:result}

The evaluation results for our leading individual systems on the validation set, along with the results for the fusion system, are detailed in~\cref{tab:result}.
% We conduct experiments using the LoRA and DoRA methods on the LLaVA-1.5-7B model.
% Compared to LoRA, DoRA necessitates training additional parameters and extends the training duration.
% However, the results indicate that the LoRA method significantly outperforms DoRA. 
While DoRA requires training more parameters and prolongs the training period compared to LoRA, our findings show that LoRA slightly outperforms DoRA.
We further fine-tune the LLaVA-NeXT-7B model using the LoRA method, which shows superior performance on both the accuracy and language metrics. 
In our extensive analysis, we integrate inference results from various individual systems, not just the ones noted as optimal in~\cref{tab:result}. For multiple-choice and yes/no questions, we employ a voting method to determine the most commonly selected answer as the final answer.
For other types of questions, we choose the answer that achieves the highest evaluation score across the relevant evaluation metrics as the final answer.
Ultimately, we achieve an optimal final score of 0.7799.

% \begin{table*}[]
% \renewcommand{\arraystretch}{1.2}
% \resizebox{\textwidth}{!}{
% \begin{tabular}{l|ccccccccc|c}
% \toprule
% \multirow{2}{*}{Systems} & \multirow{2}{*}{Accuracy$\uparrow$} & \multirow{2}{*}{ChatGPT$\uparrow$} & \multicolumn{5}{c}{Language$\uparrow$} & \multicolumn{1}{c}{} & \multirow{2}{*}{Match$\uparrow$} & \multirow{2}{*}{Final Score$\uparrow$} \\
%  & \multicolumn{1}{c}{\textbf{}} & \multicolumn{1}{c}{} & \multicolumn{1}{c}{Bleu\_1} & \multicolumn{1}{c}{Bleu\_2} & \multicolumn{1}{c}{Bleu\_3} & \multicolumn{1}{c}{Bleu\_4} & \multicolumn{1}{c}{ROUGE\_L} & \multicolumn{1}{c}{CIDEr} & \multicolumn{1}{c|}{} & \multicolumn{1}{c}{\textbf{}} \\ 
%  \midrule
% LLaVA-1.5+LoRA & 0.7735 & 64.6930 & 0.9410 & 0.9210 & 0.9006 & 0.8796 & 0.9376 & 7.8761 & 92.7437 & 0.7747 \\
% LLaVA-1.5+DoRA & 0.7212 & 64.6028 & 0.8850 & 0.8282 & 0.7789 & 0.7334 & 0.8754 & 5.4098 & 92.2840 & 0.7354 \\
% LLaVA-NeXT+LoRA & \textbf{0.7861} & 64.5203 &  \textbf{0.9411} & \textbf{0.9212} & \textbf{0.9008} & \textbf{0.8798} & \textbf{0.9377} & \textbf{7.8806} & 93.7031  & 0.7785 \\ 
% \midrule
% Fusion & \textbf{0.7861} & \textbf{64.8120} & \textbf{0.9411} & \textbf{0.9212} & \textbf{0.9008} & \textbf{0.8798} & \textbf{0.9377} & \textbf{7.8806} & \textbf{93.8406} & \textbf{0.7799} \\ 
% \bottomrule
% \end{tabular}
% }
% \caption{The table displays the best results of our individual systems on the evaluation dataset, as well as the best result of the fusion system.}
% \label{tab:result}
% \end{table*}

\section{Conclusion}
\label{sec:conclusion}
This paper outlines our contributions to the track on Driving with Language of CVPR 2024 Autonomous Grand Challenge. 
We develop vision language model systems, exclusively training them on the DriveLM-nuScenes dataset. 
Our systems are based on LLaVA models, which we enhance using the LoRA and DoRA fine-tuning techniques. 
% To improve the quality of our training data, we undertake thorough data cleaning and deduplication. 
We also integrate depth information from open-source depth estimation models into both training and inference phases. 
% For answering multiple-choice and yes/no questions during inference, we adopt a chain-of-thought reasoning method to improve accuracy. 
These efforts culminate in a notable score on the validation set leaderboard.
{
    \small
    \bibliographystyle{ieeenat_fullname}
    \bibliography{main}
}


\end{document}